\documentclass[conference]{IEEEtran}
\IEEEoverridecommandlockouts
\usepackage{cite}
\usepackage{amsmath,amssymb,amsfonts}
\usepackage{algorithmic}
\usepackage{hyperref}
\usepackage{graphicx}
\usepackage{textcomp}
\usepackage{xcolor}
\usepackage{comment}
\def\BibTeX{{\rm B\kern-.05em{\sc i\kern-.025em b}\kern-.08em
    T\kern-.1667em\lower.7ex\hbox{E}\kern-.125emX}}
\begin{document}

\title{GRAPH-GRPO-LEX: Contract Graph Modeling and Reinforcement Learning with Group Relative Policy Optimization}

\author{Moriya Dechtiar$^{*1}$, Daniel Martin Katz$^{2,3,4,5}$, Mari Sundaresan$^{6}$, Sylvain Jaume$^{7}$, Hongming Wang$^{1}$%
\thanks{$^{*}$Corresponding author: moriya.dechtiar@alumni.harvard.edu}%
\thanks{$^{1}$Harvard University, USA}%
\thanks{$^{2}$Illinois Tech - Chicago Kent College of Law, USA}%
\thanks{$^{3}$CLTDS, Bucerius Law School, Germany}%
\thanks{$^{4}$Yong Pung How School of Law, Singapore Management University}%
\thanks{$^{5}$CodeX - The Stanford Center for Legal Informatics, Stanford University}%
\thanks{$^{6}$Georgetown University, USA}%
\thanks{$^{7}$Massachusetts Institute of Technology, USA}%
}

\maketitle

\begin{abstract}
Contracts are complex documents featuring detailed formal structures, explicit and implicit dependencies and rich semantic content. Given these document properties, contract drafting and manual examination of contracts have proven to be both arduous and susceptible to errors. This work aims to simplify and automate the task of contract review and analysis using a novel framework for transforming legal contracts into structured semantic graphs, enabling computational analysis and data-driven insights. We introduce a detailed ontology mapping core legal contract elements to their graph-theoretic equivalents of nodes and edges. We then present a reinforcement learning based Large Language Model (LLM) framework for segmentation and extraction of entities and relationships from contracts. Our method, GRAPH-GRPO-LEX, incorporates both LLMs and reinforcement learning with group relative policy optimization (GRPO). By applying a carefully drafted reward function of graph metrics, we demonstrate the ability to automatically identify direct relationships between clauses, and even uncover hidden dependencies. Our introduction of the gated GRPO approach shows a strong learning signal and can move contract analysis from a linear, manual reading process to an easily visualized graph.  This allows for a more dynamic analysis, including building the groundwork for contract linting similar to what is now practiced in software engineering.  
\end{abstract}


\section{Introduction}
Contracts are essential to both business and society as they help foster economic flourishing by facilitating reliable transactions, partnerships and the orderly functioning of commercial relationships.  Given their clear societal and commercial utility, contracts are often very complex documents that incorporate intricate clauses, conditions, and contingencies designed to address potential disputes, liabilities, and obligations of the respective parties.  Many organizations in both the public and private sector expend significant resources conducting manual review of such agreements.  Such review is not only inherently time-consuming but is also susceptible to human error. 

This underscores the potential opportunity for using natural language processing (NLP) to help automate a range of contract oriented tasks including both contract generation as well as review and analysis \cite{b1}. Although large language models (LLMs) have recently demonstrated notable capabilities in processing complex legal knowledge \cite{b2}, their application in highly specialized legal tasks still requires careful engineering to mitigate potential limitations such as hallucinations. Retrieval-augmented generation (RAG) frameworks, which integrate LLMs with external knowledge sources, offer one promising approach to bolster factual grounding and domain relevance. Within this paradigm, knowledge graphs (KGs) are emerging as a critical enabling technology, providing a structured and semantically rich representation of the intricate relationships inherent in contractual documents. 

Another promising approach is reinforcement learning, which has been widely demonstrated to significantly improve model performance, accuracy, and alignment \cite{b3}. A recently introduced method of reinforcement learning, group relative policy optimization (GRPO), has shown tremendous performance improvement in the training of the R1 DeepSeek model \cite{b4} \cite{b5}.  Here we explore its effectiveness for the task of contract graph building. 

Knowledge construction and representation of contracts have been active areas of research \cite{b6} \cite{b7} \cite{b8}. A notable development for construction contracts is the Nested Contract Knowledge Graph (NCKG) proposed by \cite{b9} and \cite{b10}. The NCKG architecture, featuring a meta-model for clause-level knowledge and an ontology for conceptual schemas including risk-level information, is designed to capture the complex, multi-layered semantics often found in these documents, moving beyond simple binary triples. Such models provide a conceptual blueprint for representing the nuanced language and structure of contracts as graphs, a core objective of our research.

In the broader legal sphere, various approaches for building knowledge graphs (KGs) from diverse legal texts have been undertaken \cite{b11} \cite{b12} \cite{b13}. These include KG construction from structured, semi-structured, and unstructured judicial data \cite{b14}.  Alternatively, other scholars focused their work on entity and relationship extraction from similar sources using BERT-based models \cite{b15}. More recently, \cite{b16} leveraged knowledge-enhanced LLMs, specifically through prefix-tuning, to construct a Chinese Legal Knowledge Graph (CLKG) with high accuracy in extracting predefined entities and relationships. These studies collectively establish useful methodologies for identifying and structuring information from complex legal documents and data.  
Further theoretical underpinnings are provided in a systematic review on formalizing contractual agreements, which highlights critical steps like structural/semantic annotation and relationship discovery and emphasizes the challenges in modeling legal concepts and their interdependencies \cite{b17}.

Our motivation for building a ``contracts-to-graphs" system also relies on the synergy between KGs and RAG systems.  Modeling structured relational knowledge, as captured in KGs, can significantly improve performance of LLMs, allowing for high performance automated contract analysis and drafting \cite{b18}.

\section{Related Work}
\subsection{Advances in Entity and Relation Extraction for KGs}

The population of KGs from contractual text hinges on accurate entity and relationship extraction. Multi-agent LLM frameworks, such as KARMA \cite{b19}, offer scalable solutions by decomposing the extraction pipeline into collaborative agents responsible for tasks like entity discovery, relation extraction, and conflict resolution. This approach, emphasizing cross-agent verification and domain-adaptive prompting, mirrors the complex, iterative nature of contract analysis. \cite{b20} further illustrate LLM-driven KG construction for financial documents using a two-tiered LLM chain and prompt engineering to generate structured outputs. Essential to the coherence of these graphs is effective co-reference resolution, a task explored by \cite{b21} using neural methods to cluster entity mentions.

The quality of these extraction processes is also heavily reliant on robust preprocessing. The challenge of identifying external cross-references in contracts was addressed by \cite{b22} using a combination of NLP, pattern recognition, and web scraping. Internally, accurate sentence boundary detection (SBD) is paramount; \cite{b23} propose specialized open-source libraries like \textsc{NuPunkt} and \textsc{CharBoundry}, tailored for legal text's unique structural elements and abbreviations. Furthermore, the significance of domain-specific tokenization for legal and financial texts has been demonstrated by \cite{b24}, whose KL3M tokenizers improve processing efficiency and preserve term meaning. Complementing this, the KL3M Data Project \cite{b25} provides a large, copyright-clean corpus, offering valuable resources for training models on legally-sound data. These preprocessing advancements are foundational to the quality of input our system will handle.

\subsection{LLM Integration, RAG and Reinforcement Learning}

Our approach explores integrating LLMs with graph structures.  Graphs can be valuable in enhancing LLM performance and accuracy when combined with context retrieved from knowledge graphs.  Creating a KG for a contract could provide not only enhancements to LLM integration, but also provide a way to identify issues or high risk components within the contract.  Innovative RAG engines like KRAGEN \cite{b26} utilize ``graph-of-thoughts" techniques, embedding KGs into vector spaces to facilitate complex reasoning by LLMs. The general challenges of RAG systems, particularly for domain-specific or time-sensitive queries in private knowledge bases \cite{b27}, are well-documented. Benchmarks from \cite{b28} highlight issues such as noise robustness and multi-document summarization, while general implementation procedures \cite{b29} and investigations into LLM reliance on external knowledge \cite{b30} provide practical guidance.

To enhance the reliability of LLM outputs, frameworks like Self-RAG \cite{b31} enable LLMs to retrieve, generate, and critique their own outputs through self-reflection, a concept particularly relevant for fact-intensive tasks like contract review.  Barnett et al \cite{b32} have identified common failure points in RAG engineering, offering practical insights for robust system design. The development of robust models for legal text analysis necessitates high-quality data and standardized evaluation. Benchmarks such as LexGLUE \cite{b33} and LegalBENCH \cite{b34} provide crucial datasets for tasks including information extraction from contracts, enabling comparative assessment. Emerging concepts such as Large Concept Models \cite{b35}, which focus on instruction-following alignment for RAG \cite{b36} may further influence how contracts are processed and understood by LLMs. Innovations in generating synthetic preference data \cite{b37} and few-shot learning for text embedders \cite{b38} also promise to improve model training and retrieval efficiency. 

Finally, new approaches in reinforcement learning are showing effective self correction and better alignment results for generative models \cite{b39}. This includes developments with gated training methods \cite{b40}.  These suggest that contract analysis and drafting with LLMs could greatly benefit from combining them with KG-RAG, using our gated GRPO contract graph construction model.

Our work builds upon these advancements, focusing on the graph-based representation of contract clause dependencies using innovative reinforcement learning methods.

\section{Data}
In this section, we describe our dataset and our rationale for selecting a specific contract type.

\subsection{Corpus description}
We used the CUAD contract collection and selected a subset of 43 contracts for semi-automated labeling. Within this subset, we selected a family of contracts to avoid overfitting to a specific type of contracts.

\subsection{Contract type selection}
For purposes of analysis, we considered a series of criteria when selecting the specific contracts to be analyzed.  The selection criteria included the difficulty band (expected edge density and structure depth), the internal variability within the contract types in the contract family and the potential business value of the respective family of contract. We considered for inclusion the following families of contracts: 
\begin{itemize}
    \item Distribution \& Channel Sales
    \item Brand \& Marketing Alliances
    \item Strategic Collaboration
    \item IP Licensing \& Tech Transfer
    \item Service Delivery \& After-Sales
    \item Manufacturing \& Supply Chain
    \item Agency-Style Representation.

\end{itemize}

Ultimately, we selected the \textit{Distribution \& Channel Sales} family as it covers contract types that govern how rights-holders get products to end-customers through third-party intermediaries (e.g. Distributor, Reseller, Supply and Franchise). 

\section{Graph Modeling and Ontology Mapping}
\subsection{Legal Graph Ontology}
When representing contracts as graphs we first need to define our building blocks, i.e. the nodes and edges of our graphs. We have examined contracts with legal professionals and have developed the following definitions:
\subsubsection{Nodes}
Nodes are the entities or nouns in the contracts. We can categorize these by the layer they belong to. Structural entities or nodes can be defined as clauses or sections of the contracts. 
Here are common node types that can be extracted from a legal contract:
\begin{itemize}
\item[i]\textbf{Clause:} This is the primary unit that builds the sections of the contract. We can list several properties that this node would have: id - this is usually detailed in the contract text, it can be a number identifier or simply a bullet point (e.g., ``3.1"), title - it can have a header to indicate the content of this clause (e.g., ``Confidentiality Obligations"), text - the entirety of the text which will end at the start of the next clause identified, and a clause level which is the position depth of the clause within the contract.

\item[ii]\textbf{Defined Term:} This type of entity is a term for which a specific meaning is assigned in the context of this contract. It can be identified by a capitalized syntax and its appearance in the ``Definitions" section of the contract if such exist. 
We can list the following properties for DefinedTerm node: term - the exact string (e.g., ``Confidential Information"),  definition - the text description which is defining the meaning of this term within the contract, definition clause: not all defined terms appear neatly in the definition section. 

\item[iii]\textbf{Party:} A legal entity or individual involved in this contract and bound by it. Not limited to the entities that are signed on the contract. 
Properties that can be usually assigned are: name (e.g., ``ABC Corp."), role - what is the purpose of this party's involvement in the contract (e.g., ``Buyer," ``Distributor") and address - physical location or residence of the entity.

\item[iv]\textbf{Obligation:} A task or duty that one of the contract parties must perform and are accountable otherwise.
Attributes: action (``Pay Invoices"), actor ($\xrightarrow{}$ Party node), deadline (``within 30 days of receipt"). This is often extracted from deontic modal verbs, e.g. ``shall" or ``must".

\item[v]\textbf{Right / Permission:} A right a party is entitled to exercise.
Attributes: action (``Audit Records"), holder ($\xrightarrow{}$ Party node), frequency (``once per calendar year"). Extracted from verbs such as ``may" or ``is entitled to".

\item[vi]\textbf{Prohibition:} A constraint on a party's actions.
Attributes: action (``Reverse-engineer the Software"), subject ($\xrightarrow{}$ Party node). Extracted from ``shall not," ``may not".

\item[vii]\textbf{Condition:} A prerequisite that triggers another clause, obligation, or right.
Attributes: trigger (``If a Force Majeure Event continues for more than 60 days..."), operator (``IF...THEN").

\item[viii]\textbf{Reference:} An external standard, law, or document.
Attributes: name (``ISO 27001"), citation (``Article 30").

\item[ix]\textbf{Value:} A specific quantum.
Attributes: type (``Currency," ``Percentage"), amount (``\$5,000,000"), unit (``USD").
\end{itemize}

\subsubsection{Edges}

Edges in our contract graph represent the relationships between the nodes and entities we listed above. Some of the relationships we can model can be derived from the structure of the contract and some are more semantic and describe how one node refers to another. 
Here are some of the edges types we can extract:
\begin{itemize}
\item\textbf{IS\_PART\_OF / CONTAINS:} Hierarchical links.
Example: Clause\_3.1 - IS\_PART\_OF - Section\_3. This builds the document's tree structure.
\item\textbf{REFERENCES:} An explicit cross-reference.
Example: Clause\_10.2 - REFERENCES - Clause\_3.1. (e.g., "A breach of Section 3.1 shall be considered a material breach.")
\item\textbf{DEFINES:} Connects a clause to the term it defines.
Example: Clause\_1.1 - DEFINES - Defined Term(Confidential Information).
\item\textbf{USES:} Connects a clause to a Defined Term it uses.
Example: Clause\_3.1 - USES - Defined Term(Confidential Information).
\item\textbf{ASSIGNS\_OBLIGATION\_TO:} Connects an Obligation node to a Party.
Example: Obligation(Pay\_Fee) - ASSIGNS\_OBLIGATION\_TO - Party(Licensee).
\item\textbf{GRANTS\_RIGHT\_TO:} Connects a Right node to a Party.
Example: Right(Terminate) - GRANTS\_RIGHT\_TO - Party(Licensor).
\item\textbf{DEPENDS\_ON:} Makes one clause's activation conditional on another.
Example: Clause(Indemnification\_Obligation) - DEPENDS\_ON - Clause(Breach\_Occurred). (e.g., "Subject to Section 8...")
\item\textbf{MODIFIES / AMENDS:} Used for amendments or addenda.
Example: Amendment\_Clause\_A - MODIFIES - Original\_Contract\_Clause\_4.2.
\item\textbf{SUPERSEDES:} Indicates that one clause overrides another.
Example: Clause(Order\_of\_Precedence) - SUPERSEDES - Clause(General\_Terms).
\item\textbf{CONTRADICTS:} (Advanced) Identifies logical conflicts, often found during analysis.
Example: Clause\_A - CONTRADICTS - Clause\_B. (e.g., one clause sets payment at Net 30, another at Net 60).
\end{itemize}

\subsubsection{Creating a Contract Linter - Graph Metrics \& Algorithms}
Our goal is not only to build a graph representation for a legal contract, but also to leverage common graph metrics to derive insights from our contract graph. The following are several common graph metrics, the way to calculate them, and the way we can interpret them in a legal context to extract meaningful insights regarding the contract the graph represents. 
Related work \cite{b1}  identified significant parallels between software engineering and legal drafting, building a contract graph allows us to take this similarity to an actionable level and create a linter for contractual agreements. 
\begin{itemize}
    \item\textbf{Graph Density:} This is the ratio between actual edges in the graph and the maximum number possible. In legal context this can measure the overall contract complexity. A high-density graph is an indication the contract is convoluted and any small change has widespread effects. This type of contract would benefit from derisking strategies and manual professional review. On the other hand a too low-density graph could indicate overly simple contract that could be missing on coverage, terms and other critical components. 
    \item\textbf{Dependency Depth:} This is the length of the longest path that exists within our contract graph. This path represents the heaviest cognitive load required in order to understand the ``deepest rabbit hole." This metric helps us quantify the risk involved with the contract and directly measures how far a reader must dive in order to get a complete picture, realizing every step in this path introduces a new risk of misinterpretation. 
    \item\textbf{Degree Centrality:} The number of edges connected to a certain node. This can help us identify key clauses in a contract. 
    \item\textbf{K-Core Decomposition:}	This contract level metric that builds on the node centrality, will identify for us the ``minimum viable read" that covers the most interconnected subgraph. This subgraph is built of clauses that are all mutually co-dependent and form the heart of the agreement. Modifying any clause in the heart of the agreement has highest probability of creating unintended consequences.
    \item\textbf{In-Degree:} The number of edges going to a clause. A clause with high in-degree is referenced often within the contract and is likely a fundamental clause with wide change impact and high risk potential if ambiguous or problematic. This would be a top priority candidate for manual legal review.
    \item\textbf{Out-Degree:} The number of outgoing edges from a given node. A high number of outgoing edges suggests this clause combines many other parts of this contract and it therefor a significant connector with wide impact. 
    \item\textbf{Orphan \& Leaf Ratios:} We have defined above the degree-in and degree-out of nodes in our graph. While these help us identify node importance, we can calculate the ratios of degree-in=0 (orphan) and degree-out=0 (leaf) cases in our graph and get a metric that measures completeness and integrity of our contract data. This is a graph level quality control metric. 
    An orphan node could be one that its activating conditions are unstated within the contract. 
    A leaf node is likely a terminal node in a logical path. These are often the ultimate consequences: a payment obligation, a termination right, a specific penalty. Mapping these is crucial for understanding the final impact of a chain of events.	
    \item\textbf{Articulation Points:} An articulation point in a graph is a node or edge that is bridging otherwise disconnected sections of the graph. Removal or modification of such component does not just alter meaning, it breaks the logical tie between sections of the contract. This could mark a ``single point of failure" and needs to be identified and looked at for poor drafting or ambiguity. A poorly worded articulation point could result in contract failure.
    \item\textbf{Definition Coverage:} This coverage identifies Defined Term leafs and orphans and audits the Contract's Internal Lexicon. 
    We will list defined terms that are unused and therefore are creating glossary bloat, as well as list undefined terms that were used but not properly defined. 
\end{itemize}

Common graph algorithms can be applied on our graph to extract further insights and allow dynamic analysis.  
\begin{itemize}
    \item\textbf{Path Finding:} Listing the path bringing us from one clause to another is extremely valuable in contract graphs. This can allow us to answer what-if analysis, consider options, and possible impacts with legal reasoning. Questions such as ``What happens if we fail to deliver on time?" can be answered by finding the path from the Obligation node to the breach clauses and cures. This makes complex consequence chains explicit and clear.
    \item\textbf{Cycle Detection:} This key graph algorithm identifies paths that start and end on the same node. By doing so, we detect logical flaws and ambiguity. A cycle often represents circular logic (e.g., ``Term A is defined by reference to Term B, and Term B is defined by reference to Term A"). This is a critical flaw in legal drafting but is very difficult to spot when dealing with long contracts 
    that count many terms and conditions.

\end{itemize}

\section{Model Architecture and Experiments}
\subsection{Ground Truth Creation: The Annotation Process}
\subsubsection{Pre-processing}
Since our focus here is to create a graph from a list of clauses, segmenting the contracts into clauses is part of our pre-processing. We have a total of 43 contracts in our selected contract family. We have used \textsc{NUPunkt} and \textsc{CharBoundery} introduced in \cite{b23} to parse our contracts into clauses.
The output for each contract is a JSON file with the following properties: contract\_id, metadata and clauses, when clauses is an array of all clauses identifying in the original contract. 

This is then broken into individual clauses to be our samples for minigraph generation content. Initially, we pursued a phased approach of creating nodes and then creating entities.  However, this approach proved unworkable as the required context doubled the size of our content for the task. Each clause is the input for the creation of our minigraph of nodes and edges that can be inferred from it. Later on, these minigraphs are assembled and de-duplicated into the full contract graph. 

\subsubsection{Data Labels}
Annotators were provided with a set of definitions to help identify the requested labels. The following types of nodes were targeted for extraction: CLAUSE, PARTY, DEFINED\_TERM, VALUE. The following types of relationships were targeted for extraction: IS\_PART\_OF, REFERENCES, USES, MENTIONS\_PARTY, DEFINES, CONTAINS. Since dependency in legal contracts is both semantic and structural, both layers are modeled in our extracted relationships.
\begin{itemize}
\item\textbf{IS\_PART\_OF:} Captures the structural dependency by connecting a child CLAUSE to its parent CLAUSE.  
\item\textbf{DEFINES:} Connects a structural CLAUSE node with the semantic DEFINED\_TERM node of that term.  
\item\textbf{USES:} Connects CLAUSE nodes using a defined term, to the semantic DEFINED\_TERM they use. 
\item\textbf{REFERENCES:} Captures direct referencing or mentioning of other CLAUSE nodes.
\item\textbf{MENTIONS\_PARTY:} Connects a structural CLAUSE node with the semantic PARTY node of the PARTY it mentions.  
\item\textbf{CONTAINS:} Connects a structural CLAUSE node with the semantic VALUE node of the value it mentions.
\end{itemize}

Our output is a JSON object containing the items extracted from the sample clause, arrays of nodes or edges.

\subsubsection{Manual Labels vs. LLM Generated}
Due to the expensive nature of manual labelling on a specialized data set of contracts, we have leveraged the alt-test method introduced in \cite{b41} to justify most of our examples being labeled by LLM. 
The test provides us with a clear and statistically significant answer to weather we can replace human annotators with LLMs for the specific task at hand. The test involves 3 human annotators and approximately 50 samples that are labeled by all of them. 
For each triple of $LLM_i$, $Human_{ji}$, $Sample_i$, we calculate the difference between the LLM results and the consensus results agreed upon by the remaining humans - $Human_{j+1}$ and $Human_{j+2}$, and compare it with the difference between the $Human_{ji}$ results and the consensus results. If the LLM result is closer, this counts as a win to the LLM. We run this on our randomly selected samples set of 50 clauses across 5 randomly selected contracts. We then calculate the p-values for each of the human annotators that were left out.
For each annotator $h_j$, we test whether the LLM aligns with the other humans at least as well as $h_j$, incorporating also a penalty $\varepsilon$; We then apply Benjamini–Yekutieli FDR with q=0.05 to obtain $\omega$. If $\omega \ge 0.5$, we can justify using LLM labels for the rest of the corpus. 
\begin{figure}[t]
\centering
\includegraphics[width=0.4\textwidth]{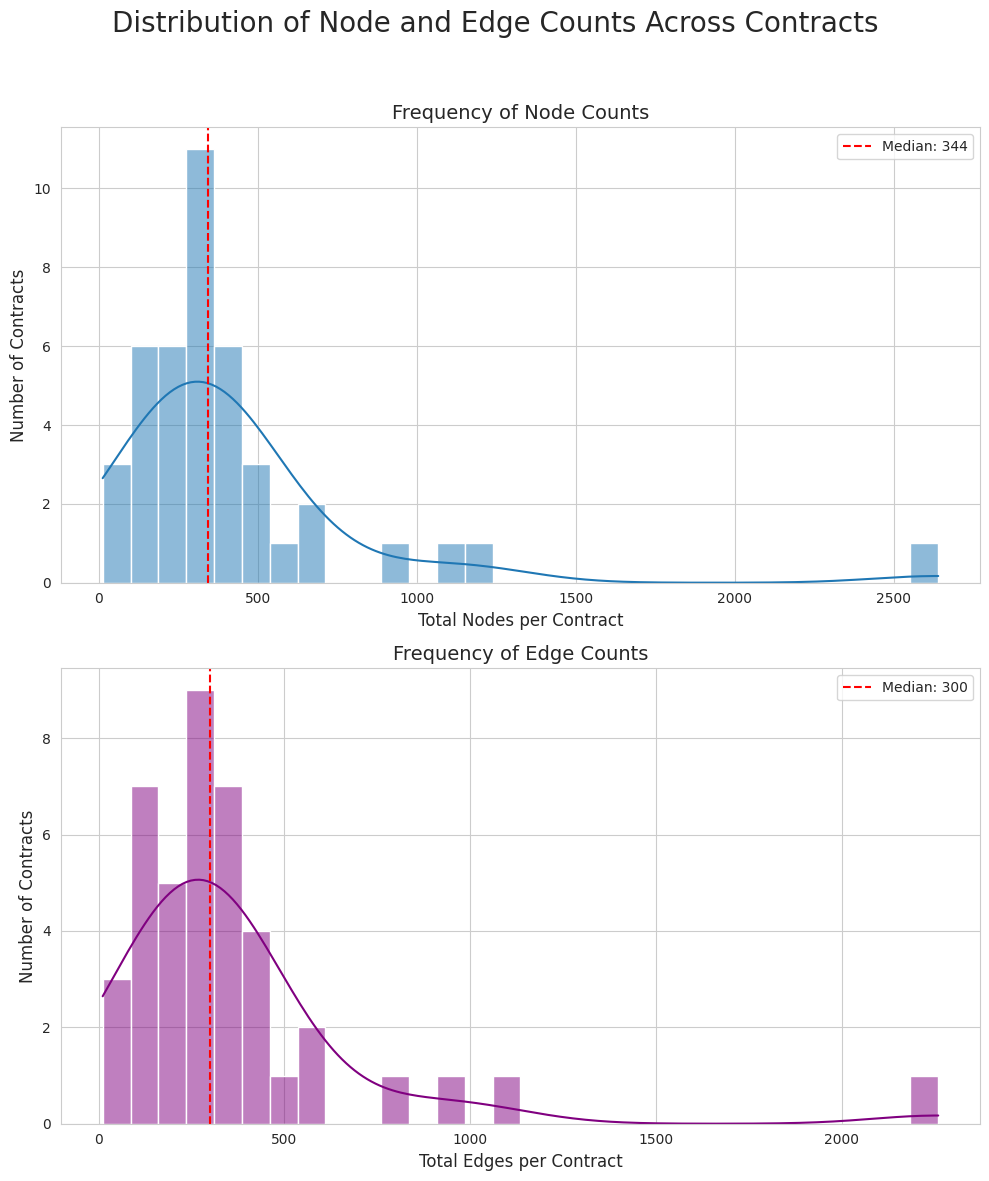}
\caption{Dataset Node Distribution Over Contracts}
\label{nodes_dist}
\end{figure}
This manually labeled data set is our held out test set, and the LLM annotated examples are used for training of our models for both nodes and edges. \\
Annotation Guidelines: To ensure consistency, both human annotators and the LLM judge were provided with a clear set of guidelines based on the ontology described in Section IV. The core instructions are explained below and can also be fully reviewed in our \href{https://github.com/moriyadechtiar/graph-grpo-lex/}{Github repository} for this work. 
\begin{figure}[t]
\centering
\includegraphics[width=0.42\textwidth]{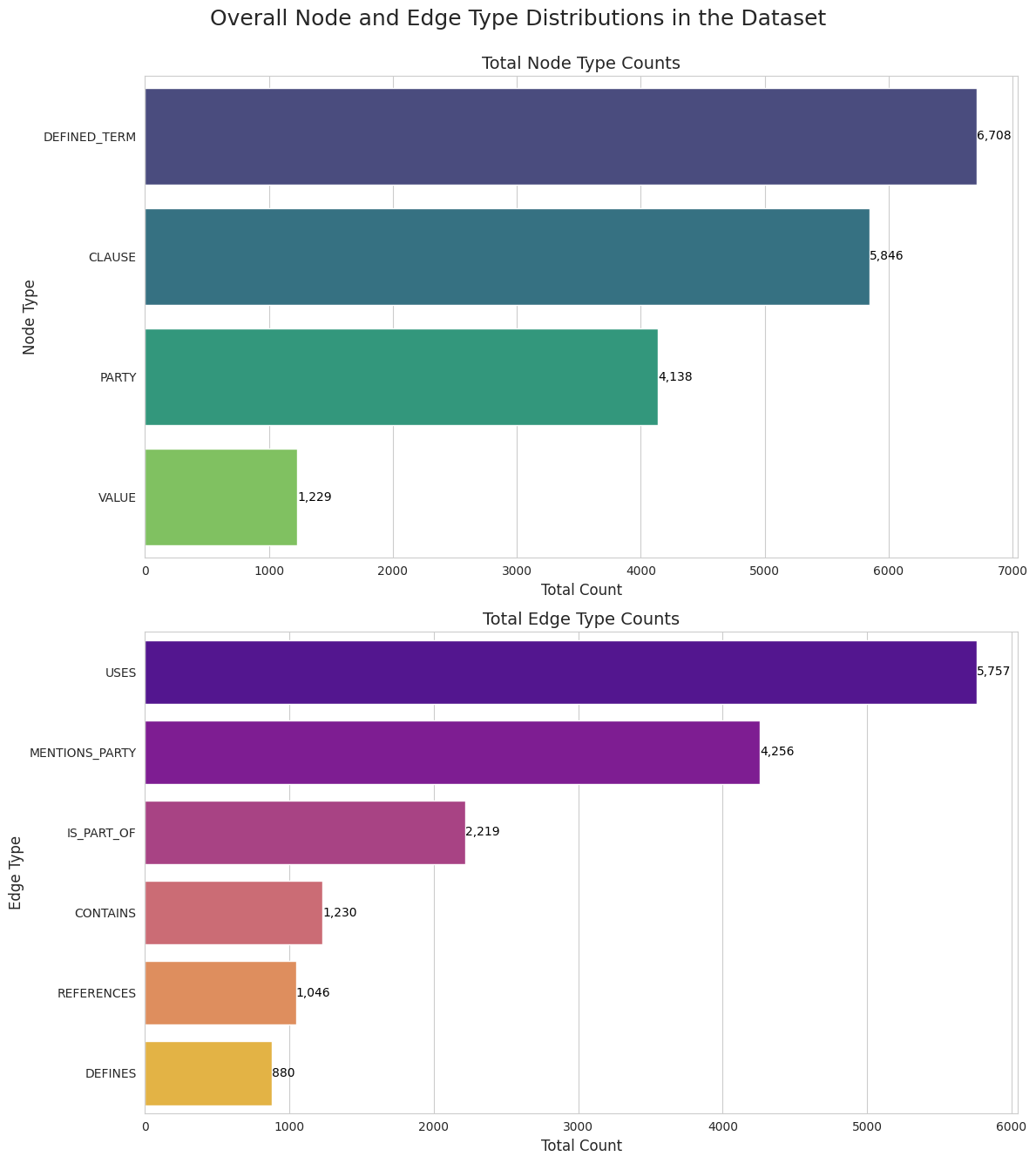}
\caption{Dataset Node and Edge Distribution by Type}
\label{all_nodes_by_type}
\end{figure}
A total of 43 contracts were parsed for annotation, containing roughly 1600 clauses. Figures \ref{nodes_dist} and \ref{all_nodes_by_type} highlight the distribution of nodes and edges within our dataset.

Alt-test results passed with the following:
\begin{equation*}
\omega = \#\{annotators~exceeded~after~FDR\} = 0.99
\end{equation*}
\begin{equation*}
Average~Advantage~Probability = 0.907
\end{equation*}

\subsubsection{Prompt Design \& Selection}
To obtain the best results under our limited compute resources, we have experimented with 4 different approaches for prompt design to do our automatic labels, when we consider factors such as style of prompt, length of prompt, structure and examples. 
\begin{itemize}
\item\textbf{Version 1. Simple NER:}
Inspired by \cite{b42}, the idea here was to focus on Named Entity Recognition task. It is then applied for identifying both nodes and relationships. It is simple, short and easy to understand, however provide little structural awareness.
This prompt failed to capture the interconnected nature of a legal clause and showed difficulty inferring the hierarchy. 
\item\textbf{Version 2. Structured \& Simple:}
This prompt is a small iterative improvement on V1 to better recognize that not all capitalized terms are defined within the current clause and introduces a way to capture this with instructions that are slightly more structured.
\item\textbf{Version 3. Structurally-Aware Graph Builder:}
This prompt is designed with a fundamental shift from treating the clause as text to treating it as a component of a document. The emphasis here is that the clause input is merely a part of a larger hierarchical document.
We provide explicit rules for the model to infer a parent and create the corresponding node and edge.
We also guide the model on how to identify other clauses that are mentioned. and how to create the corresponding CLAUSE nodes and REFERENCES edges.
On-the-Fly Node Creation: It empowers the model to create nodes for parents and referenced clauses, even though they are not the primary input.
However, the rules in this prompt are complex and presented in a dense block, which introduces a risk to confuse the model, especially with noisy clauses.
\item\textbf{Version 4. Step-by-Step Guided Graph Builder:}
This prompt focuses on improving the clarity and robustness of the instructions to increase the probability of success. 
It provides a reasoning process and mental model for the LLM to follow.
It also adds final clarifications and guardrails to check before completion. This is the longest and most complex prompt in our toolbox, and it is the most likely to introduce a complexity wall.
\end{itemize}
Evaluation: V1 and V2 underperformed and failed on even basic entries such as defined terms clauses. 
They observed to correctly create a node for the clause itself but failed to identify its relationship to the parent article, resulting in a disconnected and largely useless graph.
Prompt V3 succeeded in correctly identifying parents, even with its imperfect reasoned instructions. 
Prompt V4 further improved on and refined V3 by adding a structured REASONING PROCESS and CRITICAL CLARIFICATIONS. These instructions explicitly told the model to mentally isolate contractual text from "noise" and reinforced critical rules, such as the mandatory typing of clause references as CLAUSE nodes, not DEFINED\_TERM nodes.
This prompt achieved the best results. 
These 4 prompts were tested with state of the art models, gpt-5, grok-4 and gemini-2.5-pro. 
OpenAI gpt-5 achieved the highest usable recall on DEFINED\_TERMs and REFERENCES, good schema and guardrails adherence
gpt-5 also extracted the most definitions and cross-refs and is therefore the safer and higher yield extractor. 

\subsection{Visualizing the Zogenix Inc. Distributor Agreement}
To provide an example of the utility of our approach, consider a single agreement from our data set - the 2019 distributorship agreement between Zogenix Inc. and Nippon Shinyaku Company Ltd.  This agreement grants Nippon Shinyaku exclusive rights to distribute Zogenix's specialized seizure medication called \textit{Fintepla} in Japan. Like many other agreements in our dataset, this complex contract supports millions of dollars of commercial activity.  Using this agreement as an example, we demonstrate the value of the graph representation and perform a calculation of our previously identified metrics to evaluate the contract and the risk it carries.
\begin{figure}[t]
\centering
\includegraphics[width=0.45\textwidth]{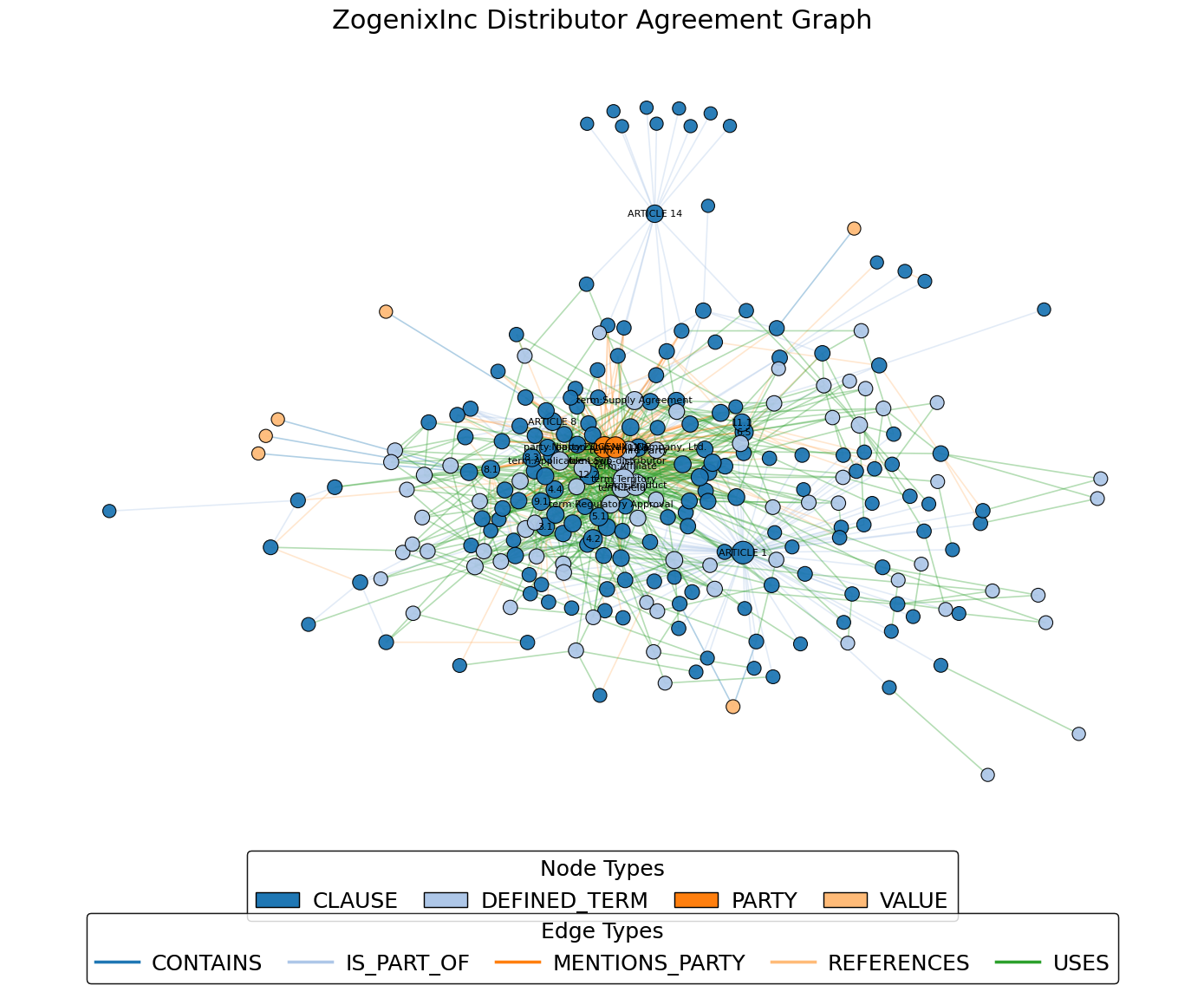}
\caption{Zogenix Inc. Contract Graph Representation}
\label{zogenixIncDist}
\end{figure}
\begin{figure}[t]
\centering
\includegraphics[width=0.45\textwidth]{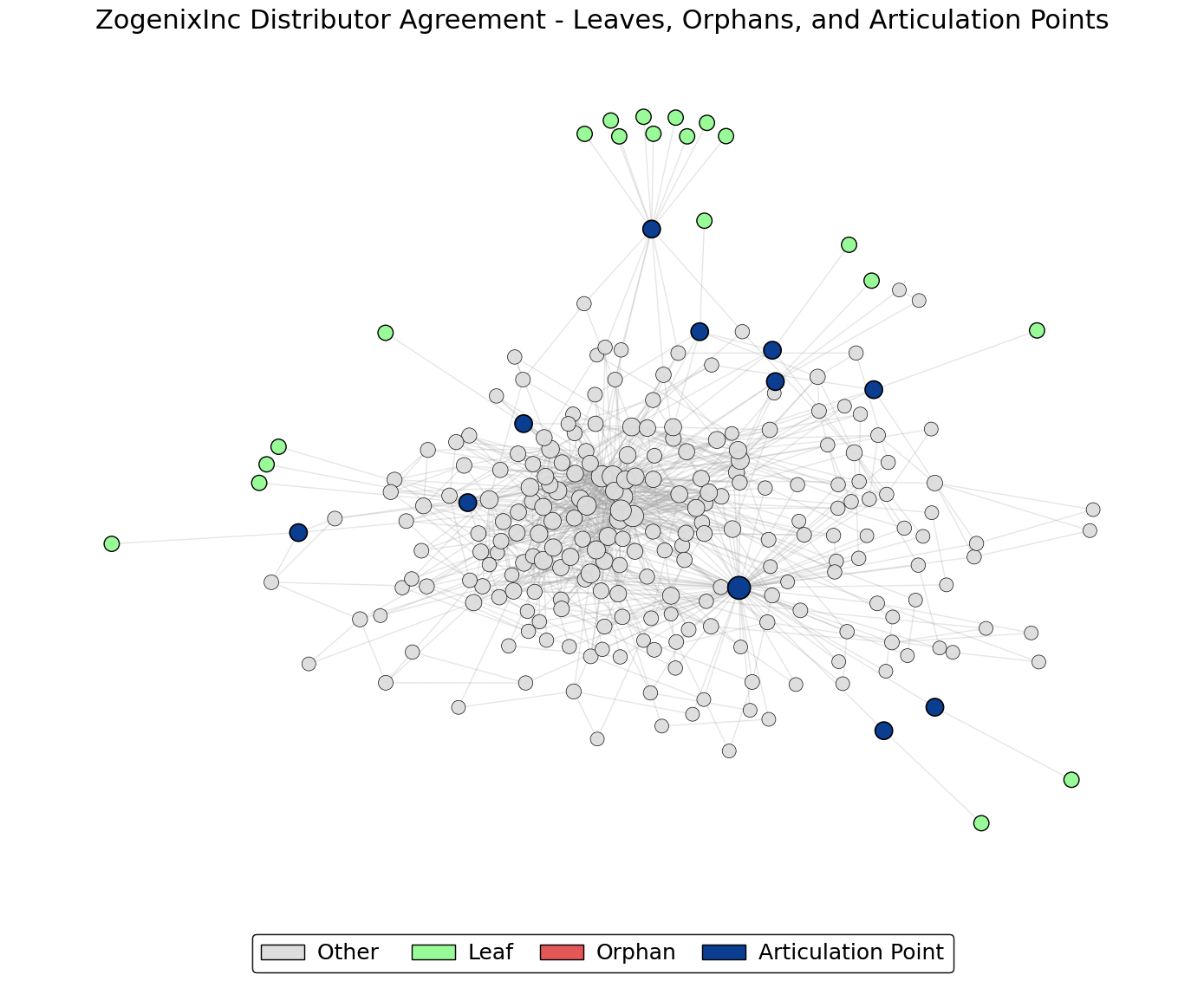}
\caption{Zogenix Inc. Contract Graph Metrics Highlights}
\label{zogenixIncHighlights}
\end{figure}
\begin{figure}[t]
\centering
\includegraphics[width=0.42\textwidth]{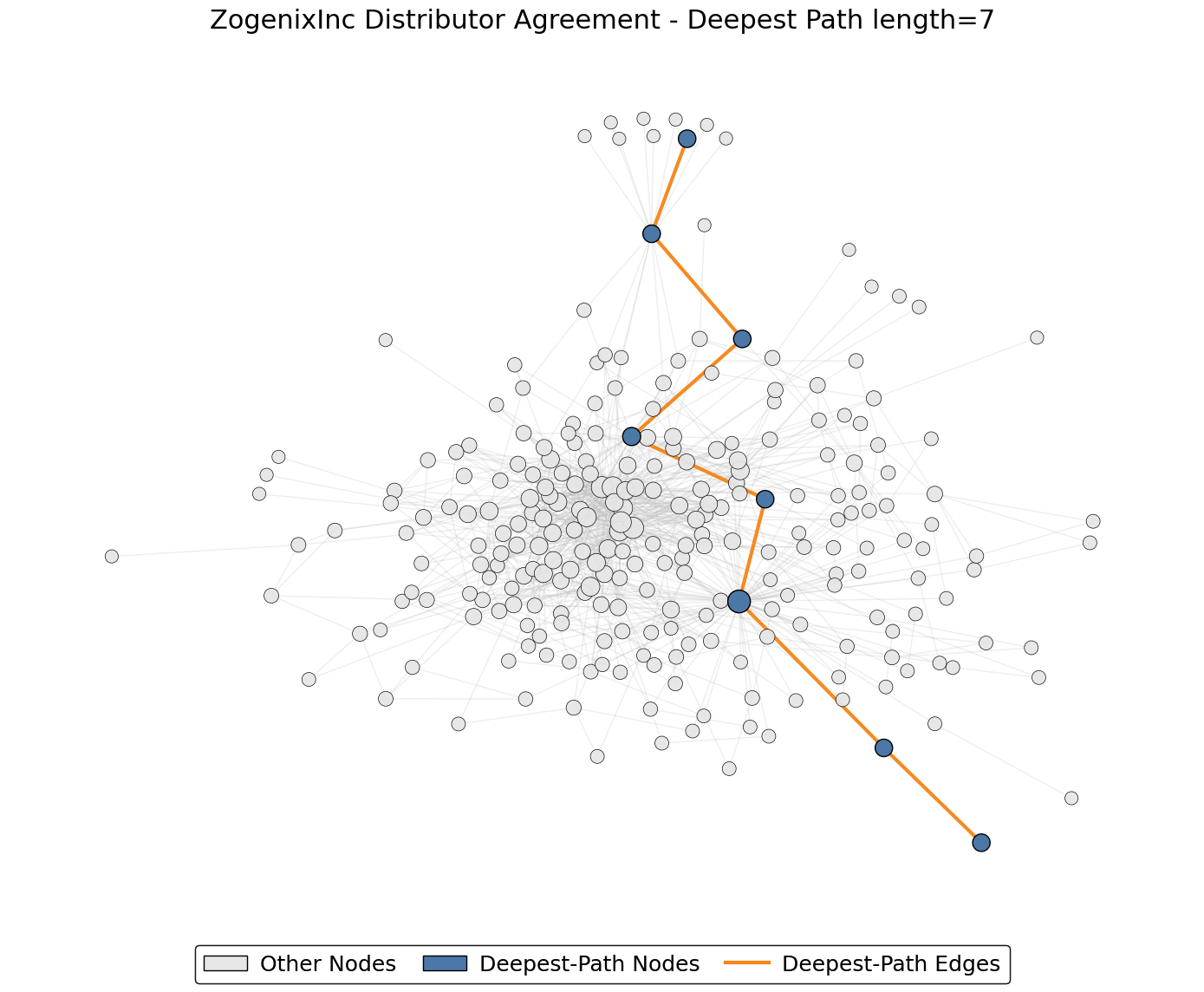}
\caption{Zogenix Inc. Contract Graph Deepest Path}
\label{zogenixIncDeepest}
\end{figure}

This agreement has 257 nodes in total (both semantic and structural) and 916 edges. Figure \ref{zogenixIncDist} offers a visualization of the graph we have created to represent this agreement based on the annotation guidelines. Figure \ref{zogenixIncDist} highlights the intricate set of relationships and entities within the contract. Revisiting our contract graph metrics, we can calculate the values for this agreement and see the following: 
\begin{itemize}
    \item 257 nodes and 916 edges: This means we are looking at a long contract which displays high connectivity.
    \item Density of 0.014 which is indication of a sparse graph relative to the maximum possible, however this number is expected when it comes to hierarchical documents. 
    \item Dependency Depth of 6 which is the longest path in this contract. The longer our deepest path is, the harder this nesting can make this agreement to review. See Figure \ref{zogenixIncDeepest} that visual this path. It is comprised of PART\_OF and REFERENCES edges. 
    \item Orphans in a legal contract might be unused terms, isolated clauses, or unlinked data. In this agreement we have 131 of these which suggests many defined items are not referenced or used.
    \item Leaves usually represent final clauses or terms that are referenced once only and this agreement has 97 which is significant however not unlikely for a hierarchical document.
    \item Orphan Ratio in this agreement is quite high and is close to 0.5 of the nodes being isolated.
    \item Leaf Ratio: 0.377 i.e. almost 40\% of our nodes are leaves.
    \item Articulation Points are nodes whose removal would split the graph into isolated components and they are critical bridges within our document. This agreement has 11 of these points holding the contract together. 
\end{itemize}

\subsection{GRAPH-GRPO-LEX: NLP Pipeline for Automated Graph Construction}

\begin{figure}[t]
\centering
\includegraphics[width=0.4\textwidth]{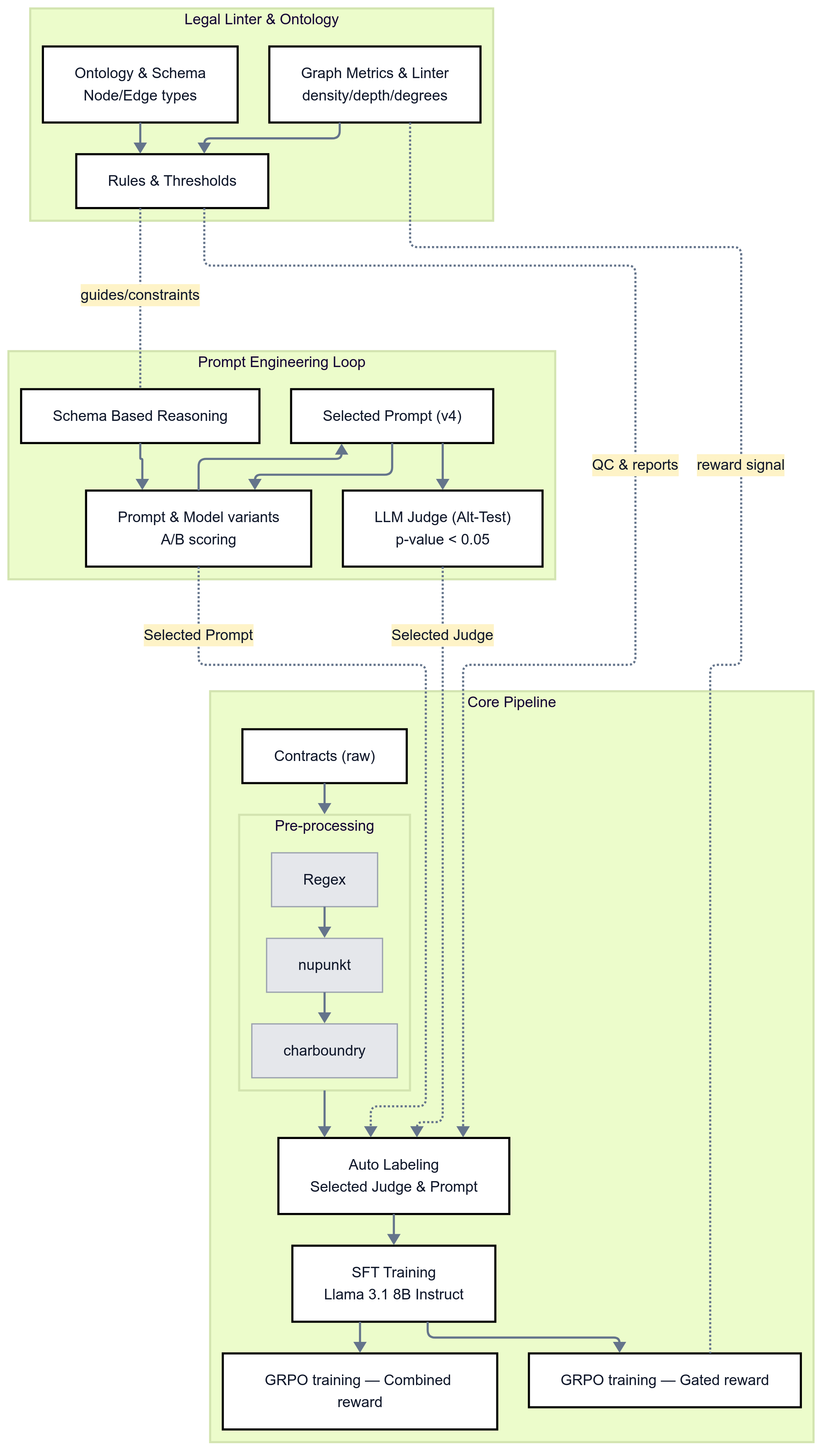}
\caption{Schematic representation of our contract graph construction pipeline, \emph{GRAPH-GRPO-LEX}}
\label{KG Construction Pipeline}
\end{figure}
We present here a  method incorporating LLMs and GRPO reinforcement learning for automated graph modeling. Our source code is available on Github.  Figure \ref{KG Construction Pipeline} provides an overview of the end-to-end Contract Graph Construction Pipeline.  This pipeline can be further broken down into particular phases as outlined below. 

\subsubsection{Legal Linter \& Ontology}
Our very first stage is developing the schema mapping legal contracts concepts into graph representation. This includes our definitions detailed earlier in this paper as well as the graph metrics translation into legal context. Those are then developed into thresholds and rules implemented both in our prompt instructions feeding our Prompt Engineering Loop, and also feeding our performance metrics calculation and the GRPO reward function development. 
\subsubsection{Prompt Engineering Loop}
This stage of our work included assembling together the parts we need in order to run our core pipeline. 
Here we A/B tested our prompts and models candidates for automated labeling. When that was completed, we could then perform our manual labelling required for the LLM-Judge statistical test which provides us with the justification to use automated labels in our core pipeline. 
\subsubsection{Core Pipeline - Pre-Processing}
All contracts are structured but they vary greatly in their style and templates. In order to create a solid and more unified structure for our models to learn from, we have included in the preprocessing stages not just cleaning of the contract, but also segmenting it and breaking it to small structural units we are referring to as clauses.  
This process started with a set of Regular Expressions (Regex) to identify patterns of clause starts and ends, header and footers and also table of contents segments in our contracts. Quickly the Regex needed to be detailed and overfit in order to produce great results to one style of contracts, but then were performing poorly when applied to a different style or contract. 
This failure brought to our attention the challenging nature of parsing contracts task. Thankfully, recent work such \cite{b23} offers more effective tools for contract segmentation. \textsc{NuPunkt} and \textsc{Charboundary} both produce the results required to get our contracts ready for graph construction. 
\subsubsection{Core Pipeline - Auto Labeling}
Each clause was paired with our selected prompt instruction, and sent to our selected large language model for inference. 
Approximately 1600 clauses were automatically labeled for our training dataset. 
\subsubsection{Core Pipeline - Supervised Fine Tuning}
In order to create a baseline for additional GRPO training, we first used SFT to train Meta-Llama-3.1-8B-Instruct with QLoRa for efficient parameter tuning, on our specific task of building minigraphs from a contract clause.
\begin{figure}[t]
\centering
\includegraphics[width=0.45\textwidth]{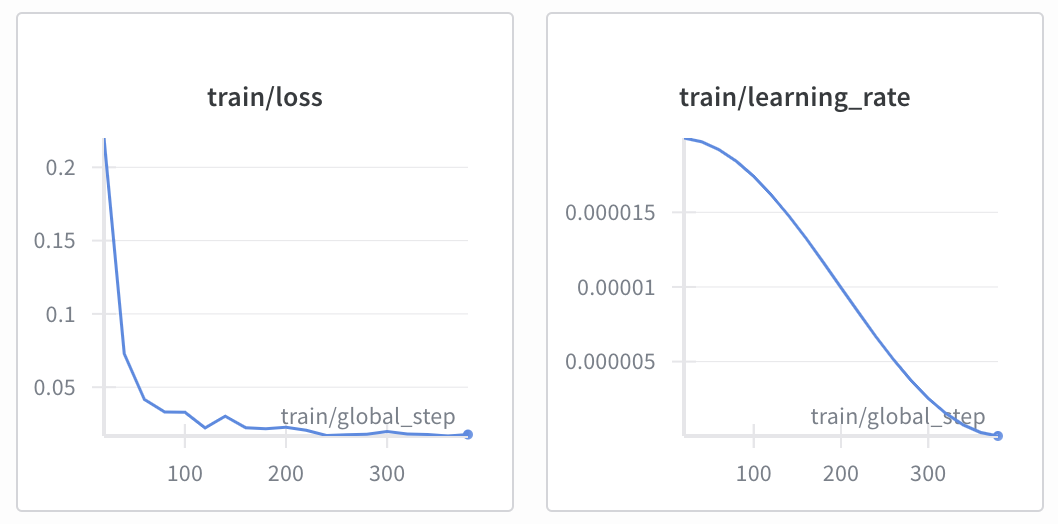}
\caption{Supervised Fine Tuning - Training}
\label{sftLoss}
\end{figure}
Figure \ref{sftLoss} shows the loss curve and learning progress during model training.
Table \ref{tab:train-eval-metrics} provides a summary of the metrics and performance of SFT.
\begin{table}[htbp]
\caption{SFT Performance: Train vs.\ Eval}
\begin{center}
\begin{tabular}{|l|c|c|}
\hline
\textbf{Metric} & \textbf{Train} & \textbf{Eval} \\
\hline
strict\_micro\_precision & 0.625 & 0.5394 \\
strict\_micro\_recall & 0.6875 & 0.8541 \\
strict\_micro\_f1 & 0.6547 & 0.6612 \\
fuzzy\_micro\_precision & 0.6292 & 0.5569 \\
fuzzy\_micro\_recall & 0.7 & 0.9166 \\
fuzzy\_micro\_f1 & 0.6627 & 0.6929 \\
invalid\_json\_rate & 0.0 & 0.0 \\
\hline
\end{tabular}
\label{tab:train-eval-metrics}
\end{center}
\end{table}

\subsubsection{Core Pipeline - GRPO Model Training}
Recent advancements introduced with DeepSeek R1 model have revealed the power of group relative policy optimization\cite{b4} \cite{b5}.  Hence we decided to test how this approach performs in the legal domain and specifically for our task of knowledge graph construction for contracts.  Our GRPO model used a basic SFT model as a baseline, and built upon of this model to create the GRPOTrainer.
We began training based on our policy guided by a reward function calculating scores for multiple generations our model is producing for each prompt. Given our finite computational resources, we selected a sample size of 4 for the generation. To select this value, we iteratively refined the components needed to produce a valid reward signal.
This process was based on the following factors:
\begin{itemize}
    \item Structure score to teach valid JSON output
    \item Strict and fuzzy F1 scores to obtain accuracy
    \item Embedding similarity for semantic score 
    \item Graph edit distance for refined accuracy
\end{itemize}

\subsubsection{Implementation Details and Hyperparameters}
Our experiments were conducted on a single NVIDIA A100 GPU (80GB VRAM), leveraging the TRL library for the GRPO implementation. SFT key hyperparameters:\\
Base Model: meta-llama/Meta-Llama-3.1-8B-Instruct\\
QLoRA settings: rank=8, lora\_alpha=16, lora\_dropout = 0.05.
Our GRPO training used a conservative learning rate of $2e-6$ with a group size of 4 generations per prompt, which yielded a stable learning signal. To balance exploration with syntactic validity, we set the decoding temperature to 0.4 with top-p sampling at 0.9, preventing collapse into non-ASCII outputs while still providing sufficient diversity for policy optimization. We applied a repetition penalty of 1.2 and used a prompt-aware JsonStopper to terminate rollouts upon forming a complete JSON object. Evaluation was performed deterministically (do\_sample=False) to ensure consistent measurement.

\section{Performance Analysis}
In this section, we describe the results of experiments conducted to evaluate performance of our GRPO training approach and detail the parameters we selected. In Figure \ref{grpo-f1-node} graphs, we show our preliminary results and provide some explanations to allow readers a deeper understanding of model training approach evolution. 

\subsection{Structured completions in high-temperature sampling} 
To obtain relative scores our model can learn from, we must sample several generated completions for each prompt in a way that will create different completions and varied scores to rank on. Using the recommended 0.6-0.7 
temperature severely impacted our model's ability to generate valid JSON structured responses. Once completions were not valid JSON, our scores were all zeros, which did not allow the model to learn. To address this issue, we adjusted the reward function adjustment to provide small reward even for partially correct JSON responses and brought the temperature down to 0.3 for sampling. 
 
\subsection{Greedy model}
To allow the model to produce even long minigraphs, we analyzed our dataset and set our maximum generated tokens based on the length of our longest dataset example. This was a value of 300 tokens.  With the greedy nature of the model, it was aiming to continue and generate all the way to the maximum value of 300 even though the correct completion should have been much shorter.  We wrapped the generated method of the model with a stopper to test for complete and valid JSON structures, forcing the model to finish at that point and avoid extra garbage tokens.

\subsection{Base tokenizer gibberish}
Our reward function has nothing in it related to the vocabulary of our generations, and since Llama is trained of a vast multilingual, varied tokens corpus, our wild initial generations included gibberish, other languages and symbols we were not expecting to see. This affected both models' ability to learn (none of these got rewarded) but also cause bottlenecks with generations taking too long, as there was nothing to trigger the JSON Stopper.  Hence, we incorporated ASCII clamp as a `LogitsProcessor` to mask at each step scores for non-ASCII tokens except a small allow-list (whitespace, quotes, braces, digits, punctuation needed for JSON).

\subsection{Computational cost}
While GRPO does not require a large dataset, it relies on additional sampling at an increased computational cost. Using an A100 GPU equipped with 80GB VRAM, we ran short training sessions with QLoRa. In our experiments, we limited the size of the sampling group to 4.

\begin{figure}[t]
\centering
\includegraphics[width=0.40\textwidth]{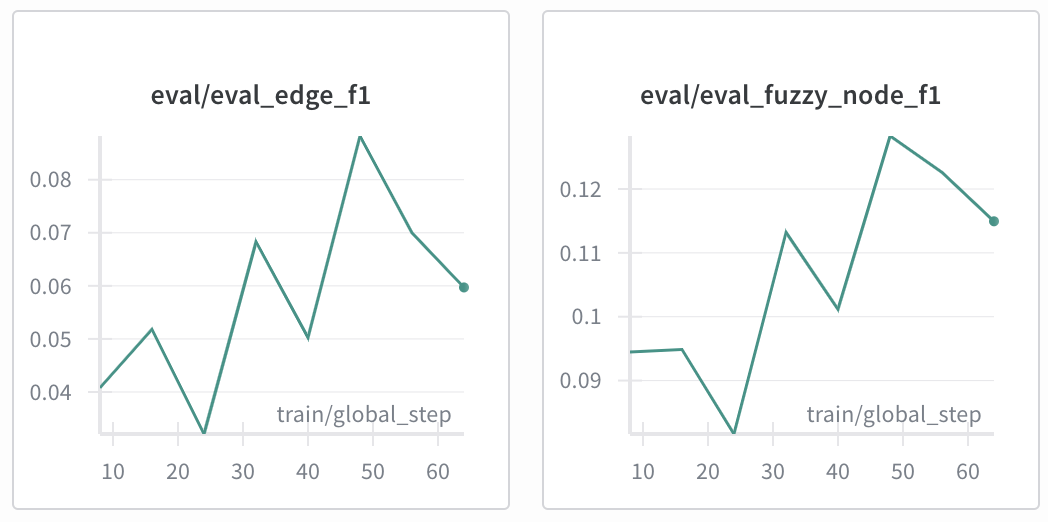}
\caption{Non Gated GRPO - Training Progress}
\label{grpo-f1-node}
\end{figure}

We experimented with different learning approaches, either teaching the skills all at once or teaching the skills one on top of the other.  To compare the performance of these competing approaches, we implemented gated access to the scores in the reward function. We ran the GRPO Trainer with the combined reward function described above and achieved a learning rate similar to the original SFT method. Note that we were able to handle a significant number of invalid JSON responses. Figure \ref{grpo-f1-node} graphs show the performance of the non-gated GRPO approach which was our naive method of computing a compound reward for All-In-One signals.

\begin{figure}[t]
\centering
\includegraphics[width=0.40\textwidth]{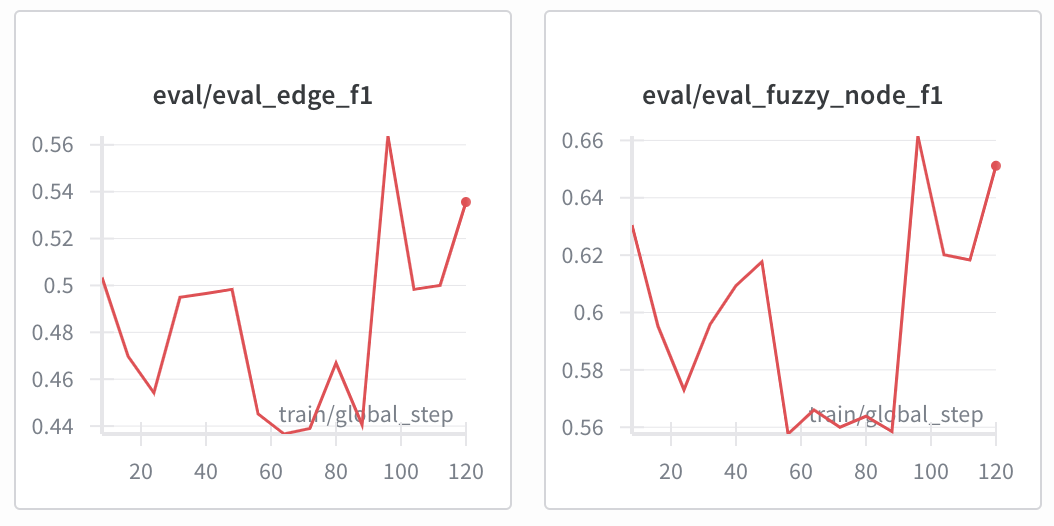}
\caption{Gated GRPO - Training Progress}
\label{gated-grpo-f1-node}
\end{figure}

These are very modest results which did not improve on our SFT training.
Hence we analyzed that the reward function was too noisy to allow significant learning and decided to teach in stages so that we open a gate to a new type of reward signal, gradually during training. 

\begin{table}[htbp]
\caption{Graph Builder Performance: GRPO vs.\ SFT Eval}
\begin{center}
\begin{tabular}{|l|c|c|}
\hline
\textbf{Metric} & \textbf{GRPO-Eval} & \textbf{SFT-Eval} \\
\hline
strict\_micro\_precision & 0.806 & 0.5394 \\
strict\_micro\_recall & 0.790 & 0.8541 \\
strict\_micro\_f1 & 0.798 & 0.6612 \\
fuzzy\_micro\_precision & 0.817 & 0.5569 \\
fuzzy\_micro\_recall & 0.802 & 0.9166 \\
fuzzy\_micro\_f1 & 0.809 & 0.6929 \\
invalid\_json\_rate & 0.02 & 0.0 \\
\hline
\end{tabular}
\label{tab:grpo-sft-eval-metrics}
\end{center}
\end{table}
Figure \ref{gated-grpo-f1-node} shows clearly that upon introduction of a new reward signal, we lose some knowledge, but then improve as we learn the current new gated reward. This staged learning approach shows 6 times better F1 scores and is a strong indication of effective learning compared to the non-gated approach. We can also see in Table~\ref{tab:grpo-sft-eval-metrics} the comparison of our final evaluation scores showing a significant improvement on our SFT model results.

\section{Conclusion}
Our method \emph{GRAPH-GRPO-LEX} operationalizes contract understanding as a graph construction and analysis task.  We introduced a reinforcement learning based LLM framework to represent a contract as a graph. Preliminary results indicate that graph-first representations make contract review more transparent and auditable while laying groundwork for reinforcement learning driven improvements in legal information extraction. Our scientific contributions include:
\begin{itemize}
    \item[(i)] a domain ontology and contract "linter" suite that translate legal structure into measurable graph properties
    \item[(ii)] a clause-centric pipeline that segments text, extracts nodes/edges, and assembles contract graphs
    \item[(iii)] an alt-test validation showing LLM labels can substitute for human annotations, enabling scale at lower cost;
    \item[(iv)] a 1600-clause dataset with a detailed case study demonstrating how density, depth, orphan/leaf ratios, and articulation points spotlight complexity and risk
    \item[(v)] a novel custom combined entities and relationship reward function for graph modeling 
    \item[(vi)] introducing gated GRPO training approach demonstrating stable reward signal and achieving significant improvement over the SFT baseline 
\end{itemize}
Overall, we have created a complete contract-to-graph system, as well as the first contract linter for automatic contract analysis and drafting.

\vspace{12pt}
\color{red}
\end{document}